# Learnable PINs: Cross-Modal Embeddings for Person Identity

Arsha Nagrani*[0000−0003−2190−9013], Samuel Albanie*[0000−0003−1732−9198], and Andrew Zisserman[0000−0002−8945−8573]

VGG, Department of Engineering Science, Oxford
{arsha,albanie,az}@robots.ox.ac.uk

**Abstract.** We propose and investigate an identity sensitive joint embedding of face and voice. Such an embedding enables cross-modal retrieval from voice to face and from face to voice.

We make the following four contributions: first, we show that the embedding can be learnt from videos of talking faces, without requiring any identity labels, using a form of cross-modal self-supervision; second, we develop a curriculum learning schedule for hard negative mining targeted to this task that is essential for learning to proceed successfully; third, we demonstrate and evaluate cross-modal retrieval for identities unseen and unheard during training over a number of scenarios and establish a benchmark for this novel task; finally, we show an application of using the joint embedding for automatically retrieving and labelling characters in TV dramas.

**Keywords:** Joint embedding, cross-modal, multi-modal, self-supervised, face recognition, speaker identification, metric learning

## 1 Introduction

Face and voice recognition, both non-invasive and easily accessible biometrics, are the tools of choice for a variety of tasks. State of the art methods for face recognition use face embeddings generated by a deep convolutional neural network [39, 41, 46] trained on a large-scale dataset of labelled faces [10, 19, 24]. A similar path for generating a voice embedding is followed in the audio community for speaker recognition [14, 33, 35, 54]. However, even though a person can be identified by their face or their voice, these two 'modes' have been treated quite independently – could they not be considered jointly?

To that end, the objective of this paper is to learn a *joint* embedding of faces and voices, and to do so using a virtually free and limitless source of unlabelled training data – videos of human speech or 'talking faces' – in an application of cross-modal self-supervision. The key idea is that a subnetwork for faces and a subnetwork for voice segments can be trained jointly to predict whether a face corresponds to a voice or not, and that training data for this task is freely available: the positives are faces and voice segments acquired from the same 'talking face' in a video, the negatives are a face and voice segment from different videos.

---

* Equal contribution

What is the motivation for learning such a joint embedding? First, a joint embedding of the modalities enables cross-modal retrieval – a person's face can retrieve face-less voice segments, and their voice can retrieve still photos and speech-less video segments. Second, this may in fact be how humans internalise identity. A highly-influential cognitive model due to the psychologists Bruce and Young (1986) [7] proposed that 'person identity nodes' or 'PINs' are a portion of associative memory holding identity-specific semantic codes that can be accessed via the face, the voice, or other modalities: and hence are entirely abstracted from the input modality.

It is worth first considering if a joint embedding is even possible. Certainly, if we task a network with learning a joint embedding then it is likely to succeed on the training data – since arbitrary associations can be learnt even from unrelated data [53]. However, if the relationship between face and voice is completely arbitrary, and the network has 'memorised' the training data then we would expect chance behaviour for cross-modal retrieval of identities that were *unseen and unheard* during training. It is unlikely that the relationship between face and voice is completely arbitrary, because we would expect some dependence between gender and the face/voice, and age and the face/voice [34]. Somewhat surprisingly, the experiments show that employing cross-modal retrieval on the joint embeddings for unseen-unheard identities achieves matches that go beyond gender and age.

In this paper we make the following four contributions. First, in Sec. 3, we propose a network architecture for jointly embedding face and voice, and a training loss for learning from unlabelled videos from YouTube. Second, in Sec. 4, we develop a method for curriculum learning that uses a single parameter to control the difficultly of the within-batch hard negatives. Scheduling the difficulty of the negatives turns out to be a crucial factor for learning the joint embedding in an unsupervised manner. Third, in Sec. 7, we evaluate the learnt embedding for unseen-unheard identities over a number of scenarios. These include using the face and voice embedding for cross-modal verification, and '1 in N' cross-modal retrieval where we beat the current state of the art [34]. Finally, in Sec. 8, we show an application of the learnt embedding to one-shot learning of identities for character labelling in a TV drama. This again evaluates the embeddings on unseen-unheard identities.

## 2  Related Work

**Cross-modal embeddings:** The relationship between visual content and audio has been researched in several different contexts, with common applications being generation, matching and retrieval [26, 29, 31]. The primary focus of this work, however, is to construct a shared representation, or joint embedding of the two modalities. While joint embeddings have been researched intensively for im-ages and text, [5, 17, 18, 28, 49], they have also started to gain traction for audio and vision [1, 4, 37, 44]. There are several ways in which this embedding may be learned—we take inspiration from a series of works that exploit audio-visual correspondence as a form of self-supervised learning [2, 38]. It is also possible to learn the embedding via cross-modal distillation [1, 4, 21] in which a trained model (the "teacher") transfers its knowledge in one modality to a second model (the "student") in another to produce aligned representations.

Of particular relevance is a recent work [3] that learns a joint embedding between visual frames and sound segments for musical instruments, singing and tools. Our problem differs from theirs in that ours is one of fine grained recognition: we must learn the subtle differences between pairs of faces or pairs of voices; whereas [3] must learn to distinguish between different types of instruments by their appearance and sound. We also note a further challenge; human speech exhibits considerable variability that results not only from *extrinsic* factors such as background chatter, music and reverberation, but also from *intrinsic* factors, which are variations in speech from the same speaker such as the lexical content of speech (the exact words being spoken), emotion and intonation [35]. A person identity-sensitive embedding must achieve invariance to both sets of factors.

**Cross-modal learning with faces and voices:** In biometrics, an active research area is the development of multimodal recognition systems which seek to make use of the *complementary* signal components of facial images and speech [8, 25], in order to achieve better performance than systems using a single modality, typically through the use of feature fusion. In contrast to these, our goal is to exploit the *redundancy* of the signal that is common to both modalities, to facilitate the task of cross-modal retrieval. Le and Odobez [30] try to instill knowledge from face embeddings to improve speaker diarisation results, however their focus is only to achieve better audio embeddings.

In our earlier work [34] we established, by using a forced matching task, that strong correlations exist between faces and voices belonging to the same identity. These occur as a consequence of cross-modal biometrics such as gender, age, nationality and others, which affect both facial appearance and the sound of the voice. This paper differs from [34] in two key aspects. First, while [34] used identity labels to train a discriminative model for matching, we approach the problem in an *unsupervised* manner, learning directly from videos without labels. Second, rather than training a model restricted to the task of matching, we instead learn a *joint* embedding between faces and voices. Unlike [34], our learnt representation is no longer limited to forced matching, but can instead be used for other tasks such as cross-modal verification and retrieval.

## 3  Learning Joint Embeddings

Our objective is to learn functions $f_\theta(x_f) : \mathbb{R}^F \to \mathbb{R}^E$ and $g_\phi(x_v) : \mathbb{R}^V \to \mathbb{R}^E$ which map faces and voices of the same identity in $\mathbb{R}^F$ and $\mathbb{R}^V$ respectively onto nearby points in a shared coordinate space $\mathbb{R}^E$. To this end, we instantiate $f_\theta(x_f)$ and $g_\phi(x_v)$ as convolutional neural networks and combine them to form a two-stream architecture comprising a face subnetwork and a voice subnetwork (see Fig. 1). To learn the parameters of $f_\theta$ and $g_\phi$, we sample a set $\mathcal{P}$ of training pairs $\{x_f, x_v\}$, each consisting of a face image $x_f$ and a speech segment $x_v$ and attach to each pair an associated label $y \in \{0,1\}$, where $y = 0$ if $x_f$ and $x_v$ belong to different identities (henceforth a negative pair) and $y = 1$ if both belong to the same identity (a positive pair). We employ a contrastive loss [12, 20] on the paired data $\{(x_{f_i}, x_{v_j}, y_{i,j})\}$, which seeks to optimise $f_\theta$ and $g_\phi$ to minimise the distance between the embeddings of positive pairs and penalises the negative pair distances for being smaller than a margin parameter $\alpha$. Concretely, the cost function is defined as:

$$\mathcal{L} = \frac{1}{|\mathcal{P}|} \sum_{(i,j) \in p} y_{i,j} D_{i,j}^2 + (1 - y_{i,j}) \max\{0, \alpha - D_{i,j}\}_+^2 \qquad (1)$$

where $(i,j) \in p$ is used to indicate $(x_{f_i}, x_{v_j}, y_{i,j}) \in \mathcal{P}$ and $D_{i,j}$ denotes the Euclidean distance between normalised embeddings, $D_{i,j} = ||\frac{f_\theta(x_{f_i})}{||f_\theta(x_{f_i})||_2} - \frac{g_\phi(x_{v_j})}{||g_\phi(x_{v_j})||_2}||_2$. Details of the architectures for each subnetwork are provided in Sec. 6.1.

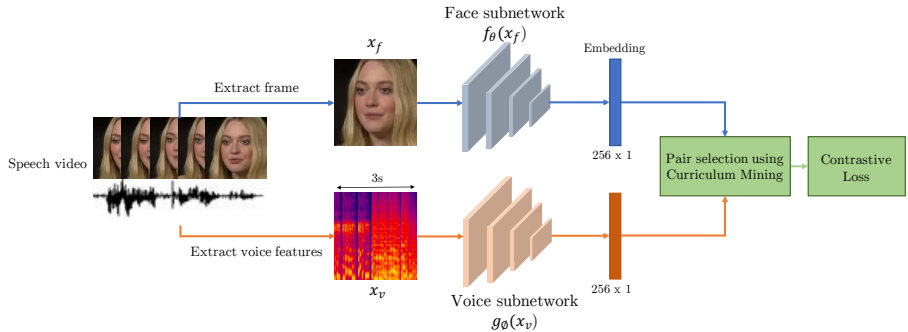

**Fig. 1.** Learning a joint embedding between faces and voices. Positive face-voice pairs are extracted from speech videos and fed into a two-stream architecture with a face subnetwork $f_\theta(x_f)$ and a voice subnetwork $g_\phi(x_v)$, each producing 256-D embeddings. A curriculum-based mining schedule is used to select appropriate negative pairs which are then trained using a contrastive loss.

### 3.1 Generating face-voice pairs

**Obtaining speaking face tracks:** In contrast to previous audio-visual self-supervised works that seek to exploit naturally synchronised data [2, 4], simply extracting audio and video frames at the same time is not sufficient to obtain pairs of faces and voice samples (of the same identity) required to train the contrastive loss described in Eqn. 1. Even for a given video tagged as content that may contain a talking human, a short sample from the associated audio may not contain any speech, and in cases when speech is present, there is no guarantee that the speaker of the audio is visible in the frame (e.g. in the case of 'reaction shots', flashbacks and dubbing of videos [36]). Furthermore, even when the face of the speaker is present there may be more than one face occupying the frame.

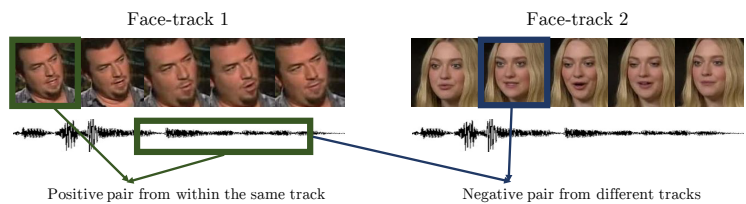

**Fig. 2.** Generating positive and negative face/voice pairs (Sec. 3.1). To prevent the embeddings from learning to encode synchronous nuisance factors, the frame for the positive face is not temporally aligned with the sequence for the voice.

We address these issues by using SyncNet [13], an unsupervised method that obtains speaking face-tracks from video automatically. SyncNet consists of a two-stream convolutional neural network which estimates the correlation between the audio track and the mouth motion of the video. This allows the video to be accurately segmented into *speaking face-tracks*—contiguous groupings of face detections from the video of the *speaker*.

**Selecting face-voice pairs**: Given a collection of speaking face-tracks, we can then construct a collection of labelled training pairs with the following simple labelling algorithm. We define face and voice segments extracted from the *same* face-track as *positive pairs* and define face and voice segments extracted from *different* face-tracks as *negative pairs* (this approach was also taken for single modality in [15]).

Since our objective is to learn embeddings that place identities together, rather than capturing synchronous, intrinsic factors (such as emotion expressions, or lexical content), we do not constrain the face associated with a positive pair to be temporally aligned with the audio. Instead it is sampled uniformly from the speaking face-track, preventing the model from learning to use synchronous clues to align the embeddings (see Fig. 2). We next describe the procedure for pair selection during training.

## 4 The Importance of Curriculum-based Mining

One of the key challenges associated with learning embeddings via contrastive losses is that as the dataset gets larger the number of possible pairs grows quadratically. In such a scenario, the network rapidly learns to correctly map the easy examples, but hard positive and negative mining [13, 22, 43, 45, 50] is often required to improve performance further. In the context of our task, a neural network of sufficient capacity quickly learns to embed faces and voices of differing genders far apart—samples from different genders then become "easy" negative pairs. Since gender forms only one of the many components that make up identity, we would like to ensure that the embeddings also learn to encode other factors. However, as we do not know the identities of the speaker face-tracks a priori, we cannot enforce sampling of gender-matched negative pairs. We tackle this issue with a hard negative mining approach that does not require knowledge of the identities during training.

When used in the unsupervised setting, hard negative selection is a somewhat delicate process, particularly when networks are trained from scratch. If the negative samples are too hard, the network will focus disproportionally on outliers, and may struggle to learn a meaningful embedding. In our setting, the hardest negatives are particularly dangerous, since they may in fact correspond to false negative labels (in which a voice and a face of the *same* identity has been sampled by chance from different speaking face-tracks)[1].

### 4.1 Controlling the difficulty of mined negatives

Standard online hard example mining (OHEM) techniques [22, 42] sample the hardest positive and negative pairs within a minibatch. However, in our setting hard positive mining may be of limited value since we do not expect the video data to exhibit significant variability within speaking face-tracks. If the hardest negative example within each mini-batch is selected, training with large batches leads to an increased risk of outliers or false negatives (i.e. pairs labelled as negatives which are actually positives), both of which will lead to poor learning dynamics. We therefore devise a simple curriculum-based mining system, which we describe next. Each mini-batch comprises $K$ randomly sampled face-tracks. For each face-track we construct a positive pair by uniformly sampling a single frame $x_f$, and uniformly sampling a three second audio segment $x_v$. This sampling procedure can be viewed as a form of simple data augmentation and makes good use of the available data, producing a set of $K$ positive face-voice pairs. Next, we treat each face input $x_f$ among the pairs as an *anchor face* and select an *appropriately hard* negative sample from within the mini-batch. This is achieved by computing the distances between its corresponding face embedding and all voice embeddings with the exception of its directly paired voice, leading to a total of $K-1$ potential negatives. The potential negatives are then ranked in descending order based on their distance to the anchor face (with the last element being the hardest negative in the batch), and the appropriate negative is chosen according to a 'negative difficulty parameter' $\tau$. This parameter simply corresponds to the percentile of the ranked negatives: $\tau = 1$ is the hardest negative, $\tau = 0.5$ the median, and $\tau = 0$ the easiest. This parameter $\tau$ can be tuned just like a learning rate. In practice, we found that a schedule that selects easier negatives during early epochs of training, and harder negatives for later epochs to be particularly effective[2]. While selecting the appropriate negative, we also ensure that the distance between the anchor face to the threshold negative is larger than the distance between the anchor face and the positive face, (following the semi-hard negative mining procedure outlined in [41]). Pseudocode for the mining procedure is provided in Appendix A and the effect of our curriculum mining procedure on training is examined in more detail in the ablation analysis (Appendix B.1), demonstrating that it plays an important role in achieving good performance.

---

[1] For a given face image and voice sampled from different speaking face-tracks, the false negative rate of the labelling diminishes as the number of identities represented in the videos grows.

[2] It is difficult to tune this parameter based on the loss alone, since a stagnating loss curve is not necessarily indicative of a lack of progress. As the network improves its performance at a certain

## 5   Dataset

We learn the joint face-voice embeddings on VoxCeleb [35], a large-scale dataset of audio-visual human speech video extracted 'in the wild' from YouTube. The dataset contains over $100,000$ segmented *speaking face-tracks* obtained using SyncNet [13] from over 20k challenging videos. The speech audio is naturally degraded with background noise, laughter, and varying room acoustics, while the face images span a range of lighting conditions, image quality and pose variations (see Fig. 5 for examples of face images present in the dataset). VoxCeleb also contains labels for the identities of the celebrities, which, we stress, are not used while learning the joint embeddings. We make use of the labels only for the purposes of analysing the learned representations – they allow us to evaluate their properties numerically and visualise their structure (e.g. Fig. 4). We use two train/test splits for the purpose of this task. The first split is provided with the dataset, and consists of disjoint videos from the same set of speakers. This can be used to evaluate data from identities seen and heard during training. We also create a second split which consists of 100 randomly selected disjoint identities for validation, and 250 disjoint identities for testing. We train the model using the intersection of the two training sets, allowing us to evaluate on both test sets, the first one for seen-heard identities, and the second for unseen-unheard identities. The statistics of the dataset are given in Table 1.

|                       | Train   | Test(S-H) | Val(US-UH) | Test(US-UH) |
|-----------------------|---------|-----------|------------|-------------|
| # speaking face-tracks | 105,751 | 4,505     | 12,734     | 30,496      |
| # identities          | 901     | 901       | 100        | 250         |

**Table 1.** Dataset statistics. Note the identity labels are not used at any point during training. SH: Seen-heard. US-UH: Unseen-unheard. The identities in the unseen-unheard test set are disjoint from those in the train set.

---

difficulty, it will be presented with more difficult pairs and continue to incur a high loss. Hence we observe the mean distance between positive pairs in a minibatch, mean distance between negative pairs in the minibatch, and mean distance between *active* pairs (those that contribute to the loss term) in the minibatch, and found that it was effective to increase $\tau$ by 10 percent every two epochs, starting from 30% up until 80%, and keeping it constant thereafter.

## 6   Experiments

We experiment with two initialisation techniques, training from scratch (where the parameters for both subnetworks are initialised randomly) and using pretrained subnetworks. In the latter formulation, both the subnetworks are initialised using weights trained for identification within a single modality. We also experiment with a teacher-student style architecture, where the face subnetwork is initialised with pretrained weights which are frozen during training (teacher) and the voice subnetwork is trained from scratch (student), however we found that this leads to a drop in performance (an analysis is provided in Appendix B.2). We use weights pretrained for identity on the VGG-face dataset for the face subnetwork, and weights pretrained for speaker identification on the VoxCeleb dataset for the voice subnetwork.

### 6.1   Network architectures and implementation details

**Face subnetwork:** The face subnetwork is implemented using the VGG-M [11] architecture, with batch norm layers [23] added after every convolutional layer. The input to the face subnetwork is an RGB image, cropped from the source frame to include only the face region and resized to $224 \times 224$. The images are augmented using random horizontal flipping, brightness and saturation jittering, but we do not extract random crops from within the face region. The final fully connected layer of the VGG-M architecture is reduced to produce a single 256-D embedding for every face input. The embeddings are then L2-normalised before being passed into the pair selection layer for negative mining (Sec. 4).

**Voice subnetwork:** The audio subnetwork is implemented using the VGG-Vox architecture [35], which is a modified version of VGG-M suitable for speaker recognition, also incorporating batch norm. The input is a short-term amplitude spectrogram, extracted from three seconds of raw audio using a 512-point FFT (following the approach in [35]), giving spectrograms of size $512 \times 300$. At train-time, the three second segment of audio is chosen randomly from the entire audio segment. Mean and variance normalisation is performed on every frequency bin of the spectrogram. Similarly to the face subnetwork, the dimensionality of the final fully connected layer is reduced to 256, and the 256-D voice embeddings are L2-normalised. At test time, the entire audio segment is evaluated using average pooling in an identical manner to [35].

The lightweight VGG-M inspired architectures described above have the benefit of computational efficiency and in practice we found that they performed reasonably well for our task. We note that either subnetwork could be replaced with a more computationally intensive trunk architecture without modification to our method.

**Training procedure:** The networks are trained on three Titan X GPUs for 50 epochs using a batch-size of 256. We use SGD with momentum (0.9), weight decay ($5E-4$) and a logarithmically decaying learning rate (initialised to $10^{-2}$ and decaying to $10^{-8}$). We experimented with different values of the margin for the contrastive loss (0.2,0.4,0.6,0.8) and found that a margin of 0.6 was optimal.

Learnable PINs         9

## 7 Evaluation

### 7.1 Cross-modal Verification

We evaluate our network on the task of *cross-modal verification*, the objective of which is to determine whether two inputs from different modalities are semantically aligned. More specifically, given a face input and a speech segment, the goal is to determine if they belong to the same identity. Since there are no available benchmarks for this task, we create two evaluation protocols for the VoxCeleb dataset, one for *seen-heard* identities and one for *unseen-unheard* identities. For each evaluation benchmark test pairs are randomly sampled, $30,496$ pairs from unseen-unheard identities and $18,020$ pairs from seen-heard identities (a description of the evaluation protocol is in Appendix C) using the identity labels provided by VoxCeleb: positives are faces and voices of the same identity, and negative pairs are from differing identities.

|  | AUC % | EER % |
|---|---|---|
| **Seen-Heard** | | |
| Random | 50.3 | 49.8 |
| Scratch | 73.8 | 34.1 |
| Pretrained | 87.0 | 21.4 |
| **Unseen-Unheard** | | |
| Random | 50.1 | 49.9 |
| Scratch | 63.5 | 39.2 |
| Pretrained | 78.5 | 29.6 |

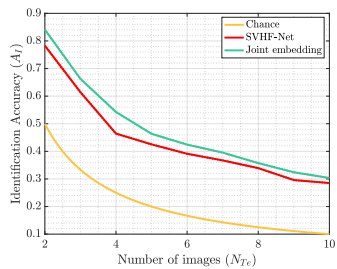

**Table 2. Cross-modal Verification:** Results are reported for an untrained model (with random weights), as well as for the two initialisations described in Sec. 6.

**Fig. 3. N-way forced matching:** We compare our joint embedding to SVHF-Net [34]. Our method comfortably beats the current state of the art for all values of N.

The results for cross-modal verification are reported in Table 2. We use standard metrics for verification, i.e area under the ROC curve (AUC) and equal error rate (EER). As can be seen from the table, the model learned from scratch performs significantly above random, even for unseen-unheard identities, providing evidence to support the hypothesis that it is, in fact, possible to learn a joint embedding for faces and voices with no explicit identity supervision. A visualisation of the embeddings is provided in Fig. 4, where we observe that the embeddings form loose groups of clusters based on identity. Initialising the model with two pretrained subnetworks brings expected performance gains and also performs surprisingly well for unseen-unheard identities, a task that is even difficult for humans to perform. Previous work has shown that on the less challenging forced matching task (selecting from two faces given a voice), human performance is around 80% [34].

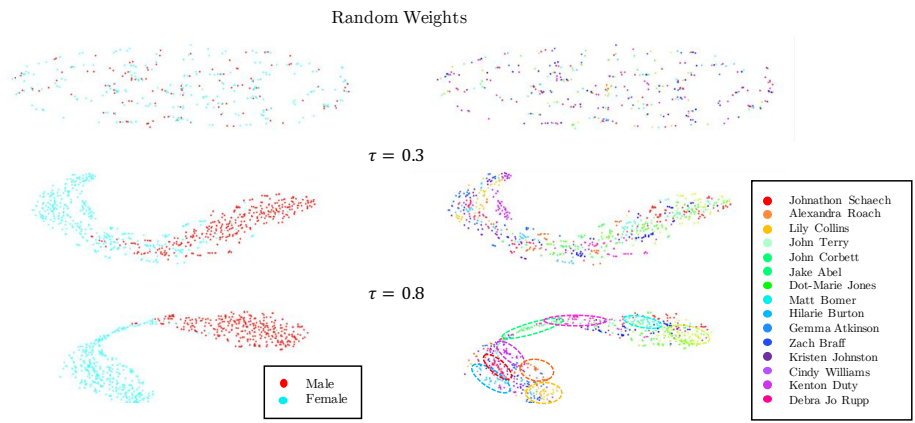

**Fig. 4.** t-SNE [32] visualisation of learnt embeddings for *faces only* from 15 identities from the VoxCeleb test set. The model is trained entirely from scratch. For visualisation purposes, embeddings are coloured with (left) gender labels and (right) identity labels (no labels were used during training). The embeddings are shown for three stages, from top to bottom; a non-trained network (random weights), a model trained with $\tau = 0.3$ and the final model trained using our curriculum learning schedule, with $\tau$ increasing from 0.3 till 0.8. Best viewed in colour.

**Effect of cross-modal biometrics:** In this section we examine the effect of specific latent properties (age, gender and nationality) which influence both face and voice. We evaluate the model by sampling negative test pairs while holding constant each of the following demographic criteria: gender (G), nationality (N) and age (A). Gender and nationality labels are obtained from Wikipedia. Since the age of a speaker could vary over different videos, we apply an age classifier [40] to the face frames (extracted at 1fps) and average the age predictions over each video (see Appendix D for more details).

| demographic criteria | random | G | N | A | GNA |
|---|---|---|---|---|---|
| unseen-unheard (AUC %) | 78.5 | 61.1 | 77.2 | 74.9 | 58.8 |
| seen-heard (AUC %) | 87.0 | 74.2 | 85.9 | 86.6 | 74.0 |

**Table 3. Analysis of cross-modal biometrics under varying demographics:** Results are reported for both seen-heard and unseen-unheard identities using AUC: Area Under Curve. Chance performance is 50%.

We find that gender is the most influential demographic factor. Studies in biology and evolutionary perception [48, 51] also show that other more subtle factors such as hormone levels during puberty affect both face morphology and voice pitch, eg. lower voice pitch correlating with a stronger jawline. However since these factors are harder to quantify, we leave this analysis for future work.

**Searching for shortcuts (bias):** As a consequence of their high modelling capacity, CNNs are notorious for learning to exploit biases that enables them to minimise the learning objective with trivial solutions (see [16] for an interesting discussion in the context of unsupervised learning). While we are careful to avoid correlations due to lexical content and emotion, there may be other low level correlations in the audio and video data that the network has learned to exploit. To probe the learned models for bias, we construct two additional evaluation sets. In both sets, negative pairs are selected following the same strategy as for the original evaluation set (they are faces and voices of different identities). However, we now sample positives pairs for the bias evaluation test sets as follows. For the first test set we sample positive pairs from the *same speaking face-track*, as opposed to sampling pairs from the same identity across all videos and speaking face-tracks (as done in our original evaluation set), and for the second test set we sample positive pairs from the *same video*. We then evaluate the performance of the model trained from scratch on the task of cross-modal verification. We obtain results that are slightly better when positive pairs are always from the same video (AUC: 74.5, EER: 33.8) vs (AUC:73.8, EER: 34.1, Table 2) on the original test set, but with minimal further improvement when they are constrained to belong to the same track (AUC: 74.6, EER: 33.6). This suggests that audio and faces taken from the same video have small additional correlations beyond possessing the same identity which the network has learned to exploit. For example, it is likely that blurry low quality videos are often accompanied by low quality audio, and that faces from professionally shot studio interviews often occur with high quality audio. While these signals are unavoidable artefacts of working with datasets collected 'in the wild', the difference in performance is slight, providing some measure of confidence that the network is relying primarily on identity to solve the task.

### 7.2   Cross-modal Retrieval with varying gallery size

The learned joint embedding also enables cross-modal retrieval. Given a single query from one modality, the goal is to retrieve all semantically matching templates from another modality (here the set of all possible templates is referred to as the *gallery set*). This can be done for both the F-V formulation (using a face to retrieve voices of the same identity) and the V-F formulation (using a voice segment to retrieve matching faces). Since there are limited baselines available for this task, we instead perform a variant of cross-modal retrieval to allow us to compare with previous work [34] (which we refer to as SVHF-Net), which represents the current state of the art for matching faces and voices. In [34], a forced matching task is used to select the *single* semantically matching template from $N$ options in another modality, and the SVHF-Net is trained directly to perform this task. Unlike this work where we learn a joint embedding, SVHF-Net consists of a concatenation layer which allows comparison of the two modalities, i.e. learnt representations in each modality are not aligned. In order to compare our method to SVHF-Net, a query set is made using all the available test samples in a particular modality. For example for the V-F formulation (used in [34]), we use all the voice segments in our unseen-unheard test set. A gallery of size N is then created for each query – a gallery consists of a single positive face and N-1 negative faces from different identities. We adopt a simple method to perform the task: the query embedding is compared directly to the embeddings of all the faces in the gallery using the Euclidean distance, and the closest embedding is chosen as the retrieved result. We compare to SVHF-Net directly on our test set, for values N = 2 to 10. A comparison of the results is given in Fig. 3.

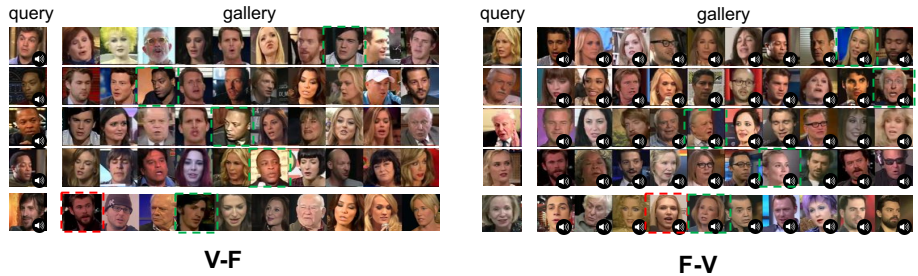

**Fig. 5.** Qualitative results for cross-modal *forced matching* (selecting the matching template from $N$ samples). We show results for $N = 10$. A query sample from one modality is shown on the left, and 10 templates from the other modality are shown on the right. For each formulation, we show four successful predictions, with the matching template highlighted in green (top four rows in each set) and one failure case (bottom row in each set) with the ground truth highlighted in green and the model prediction in red. Best viewed zoomed in and in colour.

We observe that learning a joint embedding and using this embedding directly to match faces and voices, outperforms previous work [34] for all values of $N$. In addition, note that in contrast to the SVHF-Net [34] which cannot be used if there is more than one matching sample in the gallery set, our joint embedding can be used directly to provide a ranking. In addition to the numerical results for the V-F formulation (this is the formulation used by [34]) we present qualitative results for both the V-F and face to voice (F-V) formulations in Fig. 5.

## 8   One-Shot Learning for TV Show Character Retrieval

One shot retrieval in TV shows is the extremely challenging task of recognising all appearances of a character in a TV show or feature film, with only a single face image as a query. This is difficult because of the significant visual variation of character appearances in a TV show caused by pose, illumination, size, expression and occlusion, which can often exceed those due to identity. Recently there has been a growing interest in the use of the audio-track to aid identification [9, 36, 47] which comes for free with multimedia videos. However, because face and voice representations are usually not aligned, in prior work the query face cannot be directly compared to the audio track, necessitating the use of complex fusion systems to combine information from both modalites.

For example, [9] use clustering on face-tracks and diarised speaker segments after a round of human annotation for both, [36] use confidence labels from one modality to provide supervsion for the other modality, and [47] fuse the outputs of a face recognition model, and a clothing model, with a GMM-based speaker model. With a joint embedding, however, the query face image can be compared directly to the audio track, leading to an extremely simple solution which we describe below.

**Method:** For this evaluation, we use the tracks and labels provided by [36] for episode 1 of the TV series 'Sherlock'. In order to demonstrate the effectiveness of using voice information as well, we use only the 336 speaking face-tracks from the episode, which are often the most difficult to classify visually due to large variations in head pose (it is extremely rare for the speaker to look directly at the camera during a conversation). We demonstrate our method on the retrieval of the two most frequently appearing characters, Sherlock and John, from among all the other 17 classes in the episode (16 principal characters and a single class for all the background characters).

A single query face is selected randomly for Sherlock and for John, and an embedding computed for the query using our face representation. Each face-track from the set of total tracks is then split into frames, and embeddings for each face detection are computed using our learned face representation, giving a 256-D vector for each face. The vectors are then averaged over all frames, leading to a single 256-D embedding for every track. Audio segments are also extracted for each track, and an embedding computed using our learned voice representation, giving a 256-D vector for each track in a similar fashion.

Because our representations are aligned, for each track, we can compare both the visual track and the audio track embeddings directly to the features of the query image, using L2 Euclidean distance. The tracks are then ranked according to this final score. We report results for 3 cases, retrieval using visual embeddings alone, retrieval using audio embeddings alone, and a simple fusion method where we take the maximum score out of the two (i.e. we pick the score of the modality that is closest in distance to the query image). Note, none of the identities in the episode are in the VoxCeleb training set, this test is for unseen-unheard identities. As can be seen from Table 4, using information from both modalities provides a slight improvement over using face or speech alone. Such a fusion method is useful for cases when one modality is a far stronger cue, e.g. when the face is too small or dark, or for extreme poses where the voice can still be clear [36]. On the other hand facial appearance scores can be higher when voice segments are corrupted with crosstalk, background effects, music, laughter, or other noise. We note that a superior fusion strategy could be applied in order to better exploit this complementary information from both modalities (e.g. an attention based strategy) and we leave this for future work.

|  | **Sherlock (AUC %)** | **John (AUC %)** |
|---|---|---|
| Face only | 35.0 | 44.6 |
| Voice only | 28.7 | 37.2 |
| Max Fusion | **37.5** | **45.4** |

**Table 4. One-shot retrieval results:** Retrieval from amongst 17 categories, 16 principal characters and 1 class for all the background characters. A higher AUC is better.

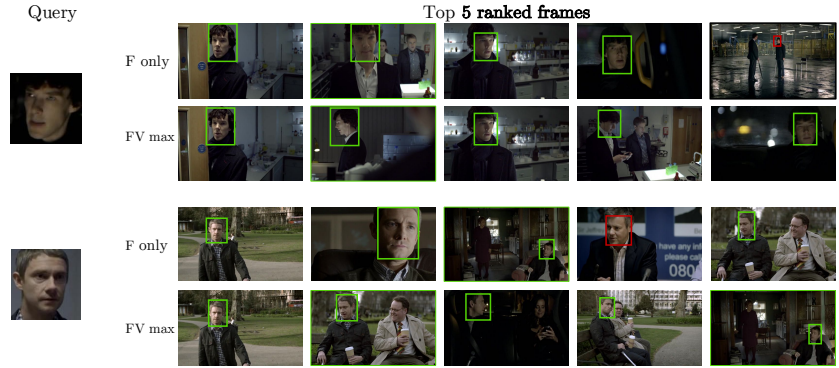

**Fig. 6.** Results of one-shot retrieval for speaking face-tracks from the TV series 'Sherlock'. A single query image and the top 5 retrieved results are shown. For each query we show tracks retrieved using only the face embeddings of the tracks (F only), and using both the face and voice embeddings (FV max). The middle frame of each retrieved track is shown. Note how FV fusion allows more profile faces to be retrieved – row 2, second and fourth frames, and row 4, third ranked frame. Face detections are green for correctly retrieved faces and red otherwise. Best viewed in colour.

## 9   Conclusion

We have demonstrated the somewhat counter-intuitive result – that face and voice can be jointly embedded and enable cross-modal retrieval for unseen and unheard identities. We have also shown an application of this joint embedding to character retrieval in TV shows. Other possible applications include biometric security, for example a face in video footage can be directly compared to an existing dataset which is in another modality, e.g. a scenario where only voice data is stored because it was obtained from telephone conversations. The joint embedding could also be used to check whether the face in a video actually matches the voice, as part of a system to detect tampering (e.g. detecting 'Deepfakes' [27]).

Identity is more than just the face. Besides voice, identity is also in a person's gait, the way the face moves when speaking (a preliminary exploration is provided in Appendix E), the way expressions form, etc. So, this work can be extended to include more cues – in accord with the original abstraction of a PIN.

**Acknowledgements.** The authors gratefully acknowledge the support of EPSRC CDT AIMS grant EP/L015897/1 and the Programme Grant Seebibyte EP/M013774/1. The authors would also like to thank Judith Albanie for helpful suggestions.

Learnable PINs    17

## A  Psuedocode for Curriculum Negative Mining

We provide pseudocode to accompany the textual description of curriculum mining described in Sec. 4 of the paper. In the algorithm below, $\tau$ defines the desired level of difficulty for the negative samples.

---
**Algorithm 1.** Curriculum Negative Mining
---
**Input:** Minibatch $B = \{x_{f_i}, x_{v_j} | i, j \in \{1 \ldots K\}\}, \tau$
**Output:** List of negatives $x_{n_i}$
1: **for** $i = 1$ to $K$ **do**  ▷ for each face embedding
2:     $D_{i,j} = \|x_{f_i} - x_{v_j}\|_2$
3:     $D_{i,j}^{Ranked}$, sort_indices $= \text{sort}(\{D_{i,j} | i \neq j\}, \text{desc})$  ▷ rank negative distances
4:     $n_\tau = \text{round}(\tau.(K-1))$  ▷ position of threshold negative
5:     $n_i = \text{argmin}_j |(D_{i,j}^{Ranked} - D_{i,i})|$  ▷ position of hardest semi-negative
6:     $p_i = \min(n_\tau, n_i)$  ▷ select final position
7:     $x_{n_i} = x_v[\text{sort\_indices}[p_i]]$
---

## B  Additional Ablation Experiments

### B.1  Effect of Curriculum Mining Procedure:

To assess the effect of the curriculum mining schedule proposed in Sec. 4 on learning subnetworks from scratch, we conduct experiments with four different negative mining techniques. The first technique involves selecting negative samples at random (and therefore not controlling the difficulty of the negatives), the second is the popular semi-hard mining technique proposed by FaceNet [41](i.e. CHNM with $\tau = 0$), the third is other fixed $\tau$ values ($\tau = 0.3, 0.5, 0.8$), and the final technique is our proposed curriculum mining schedule. Results are given in Table 5.

| mining strategy | random | $\tau = 0$ [41] | $\tau = 0.3$ | $\tau = 0.5$ | $\tau = 0.8$ | CHNM |
|---|---|---|---|---|---|---|
| AUC (%) | 50.2 | 51.2 | 51.9 | 55.8 | 59.4 | **73.8** |

**Table 5.** Comparison of different mining procedures for models containing a subnetwork trained from scratch. Results are reported for the task of cross-modal verification on seen-heard identities, using AUC %, as reported in Table 2. Chance is 50%.

Using the random mining strategy, we found the results to be similar to chance. This suggests that a "collapse of training" has occurred (a phrase coined by [22]), in which training has fallen into a local minimum that assigns the distance between positive and negatives pairs to be equal, and thereby avoids solving the task. Using semi-hard negative mining produces similar results, suggesting that the negatives are too difficult for the model trained from scratch, forcing training to collapse. These results indicate the importance of starting the training process with easy pairs to enable the model to learn useful representations. For fixed $\tau$ (0.5 and 0.8) we find that the model only learns to correctly classify easy pairs (different genders). We confirm this by testing the models on test pairs where the negatives share the same gender, and obtain a performance similar to chance (51.2%, 50.4%). We additionally compared the performance of mining strategies when initialising with two pretrained subnetworks. We found that the performance of semi-hard negative mining and curriculum mining were similar (86.4 vs 87.0), suggesting that the benefit of curriculum mining (with the schedule suggested in the paper) lies primarily in helping models trained from scratch avoid being trapped in poor local minima, rather than improving learning for existing representations. We note, however, that since curriculum learning is a direct generalisation of semi-hard negative mining (achieved by fixing the difficulty parameter $\tau$ to zero), that although it was not the primary focus of this work, it may be possible to find curriculum schedules that work more effectively for the pretrained case.

### B.2   Training Strategy

We also perform an experiment with forms of indirect supervision in the form of single modality pretraining. Inspired by the approach taken in SoundNet [4], we conduct an experiment in which the subnetwork for one modality acts as the 'teacher' and the subnetwork for the other modality is the 'student'. We use a pretrained face subnetwork with frozen weights, and then train the voice subnetwork from scratch in an attempt to 'anchor' the embeddings in the face identity space. We find, however that the results were worse than training from scratch. We note that similar findings were reported in [2]. Moreover, recent work has shown that the teacher networks trained for face verification typically encode factors beyond identity such as pose [52]. Consequently, we hypothesise that the student may be expending capacity on predicting features that cannot feasibly be learned from an audio signal.

### B.3   Training Loss

While we opt to use the contrastive loss [12, 20] in our experiments, we note that [3] found a modification of the binary softmax loss to be particularly effective for learning a joint embedding between images and audio of instrument classes from scratch. We ran an experiment using their method (for our model trained entirely from scratch), and found that it did not provide a significant difference to the results, but instead worsened results slightly (AUC: 72.1% vs 73.8% using contrastive loss). Another option which has proven popular for learning face embeddings is the triplet loss [41]. In this work, for the sake of simplicity we restricted our attention to pairwise losses to avoid asymmetries introduced by sampling triplets across modalities (three samples from only two modalities), however we note that this would make an interesting extension for future work.

## C    Evaluation Protocol for Cross-Modal Verification

The number of face images and audio segments used during training can be seen in table 6. The number of data samples in the unseen-unheard test set (disjoint identities) is much larger than the seen-heard test set (as the majority of samples from these identities are used for training). For each audio sample in the unseen-unheard test set, a positive face image (same identity) or negative image (different identity) is chosen at random from within the same test identities. Note that the positive face is selected from across all available facetracks and videos. (Since the network was trained in an unsupervised manner, this was not the exact condition under which positive pairs were sampled during training. Positive pairs were always sampled from within the same facetrack.) This gives a total of $30,496$ evaluation pairs. For every audio sample in the seen-heard test set, four random face images are selected from within the same test set (2 positive and 2 negative). This gives a total of $18,020$ evaluation pairs. The pair sampling is hence class-balanced (half positive, half negative).

|                    | Train   | Test(S-H) | Val(US-UH) | Test(US-UH) |
|--------------------|---------|-----------|------------|-------------|
| # faces            | 829,862 | 101,348   | 45,482     | 240,866     |
| # audio segments   | 105,751 | 4,505     | 12,734     | 30,496      |
| # evaluation pairs |         | 18,020    |            | 30,496      |

**Table 6.** Number of face images and voice segments used during training and test. S-H: seen-heard, US-UH: unseen-unheard.

## D    Obtaining Age labels for the VoxCeleb dataset

While gender and nationality labels can be obtained using the identities alone, this is not the case for age - since the VoxCeleb dataset consists of YouTube videos uploaded at various times, age can vary from one video to the next, even for the same identity. We extract frames at 1fps, and run a visual age classifier [40] on each face detection per frame. We replace the pretrained VGG network used in the original paper [40] with a Wide Residual Network (WideResNet); adding two classification layers (for age and gender estimation) on top of the standard WideResNet architecture. This classifier is trained on the IMDB-WIKI dataset. We make the assumption that the age of the speaker does not vary in a single video (this could be erroneous in the case of flashback videos and other mashups, however a quick manual inspection showed us that such videos are in the minority) and average age predictions over a video in order to get a single age label per video. We then define 5 age groups, $< 20$ years, $20 - 30$ years, $30-40$ years, $40-50$ years, and $50+$ years. The number of facetracks in each group can be seen in Table 7.

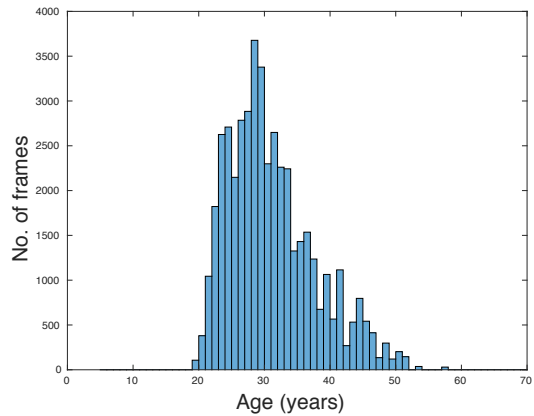

**Fig. 7.** Distribution of ages in the voxceleb test set: Age labels are obtained for each frame automatically using a visual age classifier [40].

| Age Bin | $< 20$ | $20-30$ | $30-40$ | $40-50$ | $50+$ |
| --- | --- | --- | --- | --- | --- |
| # facetracks | 374 | 16391 | 10624 | 2488 | 619 |

**Table 7.** The facetracks in the VoxCeleb dataset are binned into 5 age groups.

## E    Adding in temporal visual information

In addition to static biometrics, there could exist other (dynamic) cross-modal biometrics, such as a person's 'manner of speaking' [34]. In an attempt to capture a person's idiosyncratic speaking style, (which we believe could be correlated with the person's voice) we represent temporal facial motion using dynamic images [6]. We modify our architecture to include an extra face stream, and dynamic and RGB embeddings are then averaged to get the final face embedding. The rest of the architecture remains the same as described in the main paper. Using this joint embedding the performance on unseen-unheard identities is 80.2%, higher than that using static images alone (78.5%) due to the additional temporal information. This could be an interesting avenue for further exploration, as more identity information from different sources can be incorporated into the joint embedding.